\documentclass[twoside,11pt]{article}

\usepackage[arxiv]{melba}

\usepackage{mwe} 

\usepackage{amsmath,amsfonts}

\usepackage{multirow}

\usepackage{subcaption}

\usepackage{rotating}

\usepackage[nomargin,inline,marginclue,draft]{fixme}
\fxsetup{status=draft}
\fxsetup{theme=color, mode=multiuser} \FXRegisterAuthor{me}{ame}{\color{red}Me}

\hypersetup{
    colorlinks=true,
    citecolor=blue,
    urlcolor=purple,
}

\melbaid{YYYY:NNN}  
\doi{https://doi.org/10.59275/j.melba.2023-AAAA}
    \melbaauthors{Pati \textit{et al}}  
\volume{2}
\firstpageno{1}  
\melbayear{YYYY}  
\datesubmitted{m1/yyyy}  
\datepublished{m2/yyyy}  

\ShortHeadings{GaNDLF-Synth: Democratizing Generative AI for Medical Imaging}{Pati \textit{et al.}}

\title{GaNDLF-Synth: A Framework to Democratize Generative AI for (Bio)Medical Imaging}

\author{\firstname Sarthak \surname Pati \orcid{0000-0003-2243-8487} \email patis@iu.edu \\  
	\addr Division of Computational Pathology, Department of Pathology and Laboratory Medicine, Indiana University School of Medicine, Indianapolis, IN, USA \\
	\addr Center for Federated Learning, Indiana University School of Medicine, Indianapolis, IN, USA \\
	\addr Medical Working Group, MLCommons, San Francisco, CA, USA
	\AND
	\name Szymon Mazurek\orcid{0009-0006-7557-0157} \email szmazurek@agh.edu.pl \\
	\addr AGH University of Krakow, Academic Computer Centre Cyfronet, Krakow, Poland
	\AND
	\name Spyridon Bakas\orcid{0000-0001-8734-6482} \email spbakas@iu.edu \\
	\addr Division of Computational Pathology, Department of Pathology and Laboratory Medicine, Indiana University School of Medicine, Indianapolis, IN, USA \\
	\addr Center for Federated Learning, Indiana University School of Medicine, Indianapolis, IN, USA \\
        \addr Indiana University Melvin and Bren Simon Comprehensive Cancer Center, Indianapolis, IN, USA \\
        \addr Department of Radiology \& Imaging Sciences, Indiana University School of Medicine, Indianapolis, IN, USA \\
        \addr Department of Biostatistics \& Health Data Science, Indiana University School of Medicine, Indianapolis, IN, USA \\
        \addr Department of Neurological Surgery, Indiana University School of Medicine, Indianapolis, IN, USA \\
        \addr Department of Computer Science, Luddy School of Informatics, Computing, and Engineering, Indiana University, Indianapolis, IN, USA 
	\addr Medical Working Group, MLCommons, San Francisco, CA, USA
}

\begin{document}

\maketitle

\begin{abstract}
    Generative Artificial Intelligence (GenAI) is a field of AI that creates new data samples from existing ones. It utilizing deep learning to overcome the scarcity and regulatory constraints of healthcare data by generating new data points that integrate seamlessly with original datasets. This paper explores the background and motivation for GenAI, and introduces the Generally Nuanced Deep Learning Framework for Synthesis (\textbf{GaNDLF-Synth}) to address a significant gap in the literature and move towards democratizing the implementation and assessment of image synthesis tasks in healthcare. GaNDLF-Synth describes a unified abstraction for various synthesis algorithms, including autoencoders, generative adversarial networks, and diffusion models. Leveraging the GANDLF-core framework, it supports diverse data modalities and distributed computing, ensuring scalability and reproducibility through extensive unit testing. The aim of GaNDLF-Synth is to lower the entry barrier for GenAI, and make it more accessible and extensible by the wider scientific community.
\end{abstract}

\begin{keywords}
	Deep Learning, Synthesis, Generative Artificial Intelligence, Generative Adversarial Networks, Diffusion, Autoencoders, Framework, GaNDLF
\end{keywords}

\section{Background}
Modern artificial intelligence (AI) techniques, specifically those based on deep learning (DL), are extremely data hungry \cite{varoquaux2022machine,baheti2022leveraging,pati2022federated}. Healthcare data is relatively scarce compared to other types of data (such as those for computer vision \cite{deng2009imagenet}) and is far more tightly regulated \cite{hipaa,gdpr,disha}. Generative artificial intelligence (GenAI) encompasses a range of AI methods, techniques, and algorithms that are crafted to understand the inherent patterns and structures within a dataset, thus enabling them to produce new data points that could seamlessly integrate with the original dataset \cite{epstein2023art,pinaya2023generative}. GenAI has come to the forefront of usage across various scientific domains, including and not limited to text-to-text, text-to-image, and image-to-image\cite{cao2023comprehensive,maerten2023paintbrush,bandi2023power,gupta2024generative,sengar2024generative}. There are a lot of opportunities for using GenAI in healthcare \cite{zhang2023generative,shokrollahi2023comprehensive}, particularly for imaging, where data, in addition to being scarce, is not readily shared across acquiring sites \cite{pati2022federated}.

There are multiple applications of GenAI in the context of medical imaging (Table~\ref{tab:genai_applications}). For example, with GenAI, it is a possibility of not using \textbf{real} patient imaging data at all and instead only using \textbf{synthesized} imaging data for downstream clinical tasks (e.g., using synthesized MRI scans for predicting patient outcomes). This provides increased privacy by alleviating concerns of models leaking training information \cite{song2019membership,pati2024privacy}. GenAI can potentially rectify the situations where there is gross data imbalance, especially for rare diseases, where positive cases are much fewer. GenAI could also potentially address the issue of class imbalance across sites in multi-institutional studies \cite{pati2022federated,pati2024privacy,pati3,pati2021federated,baid2022federated,foley2022openfl,sheller2020federated}. GenAI models can be powerful in terms of learning data distribution in an unsupervised way—such trained architectures can be transferred as feature extractors for other problems, such as classification \cite{lin2017marta,shrivastava2017learning}. GenAI models can be used as enhancers of data quality, as they can be trained to perform denoising, reconstruction, or super-resolution tasks \cite{frangi2018simulation}. With proper training, there exists a possibility to train a model enabling the generation of different modalities based on some base examples, such as generating PET images (which are ionizing) from MR images (which are non-ionizing) \cite{dayarathna2023deep}. While such translations are difficult, they can possibly serve as an interesting approximation tool, and they have an added benefit reducing additional radiation to patients.

\begin{table}[ht]
\caption{Potential Applications and Benefits of GenAI in Medical Imaging}
\label{tab:genai_applications}
\resizebox{\textwidth}{!}{
\begin{tabular}{|l|l|}
\hline
\textbf{Aspect}                                                     & \textbf{Description}                                                                                                                                                                                                                                                                                                                                                               \\ \hline
\begin{tabular}[c]{@{}l@{}}Data \\ Imbalance\end{tabular}           & \begin{tabular}[c]{@{}l@{}}GenAI can address gross data imbalance, especially for rare diseases, by \\ generating more positive cases. It can also handle class imbalance across \\ sites in multi-institutional studies \\ \cite{sheller2020federated,pati2022federated}.\end{tabular} \\ \hline
\begin{tabular}[c]{@{}l@{}}Unsupervised \\ Learning\end{tabular}    & \begin{tabular}[c]{@{}l@{}}GenAI models can learn data distribution in an unsupervised manner and \\ be transferred as feature extractors for other tasks, such as \\classification \cite{lin2017marta,shrivastava2017learning}.\end{tabular}                                                                                                                       \\ \hline
\begin{tabular}[c]{@{}l@{}}Data Quality \\ Enhancement\end{tabular} & \begin{tabular}[c]{@{}l@{}}GenAI models can enhance data quality by performing tasks like denoising, \\ reconstruction, or super-resolution \cite{frangi2018simulation}.\end{tabular}                                                                                                                                                                             \\ \hline
\begin{tabular}[c]{@{}l@{}}Modality \\ Generation\end{tabular}      & \begin{tabular}[c]{@{}l@{}}With proper training, GenAI models can generate different imaging modalities \\ from base examples, such as generating PET images from \\MR \cite{dayarathna2023deep}, thus, serving as an approximation tool.\end{tabular}                                                                                                           \\ \hline
Privacy                                                             & \begin{tabular}[c]{@{}l@{}}GenAI can enable the use of synthesized data instead of real patient data,\\ which can increase privacy by reducing the risk of models leaking training \\ information \cite{song2019membership,pati2024privacy}.\end{tabular}                                                                                                       \\ \hline
\end{tabular}
}
\end{table}

\begin{figure}
    \centering
    \includegraphics[width=0.9\linewidth]{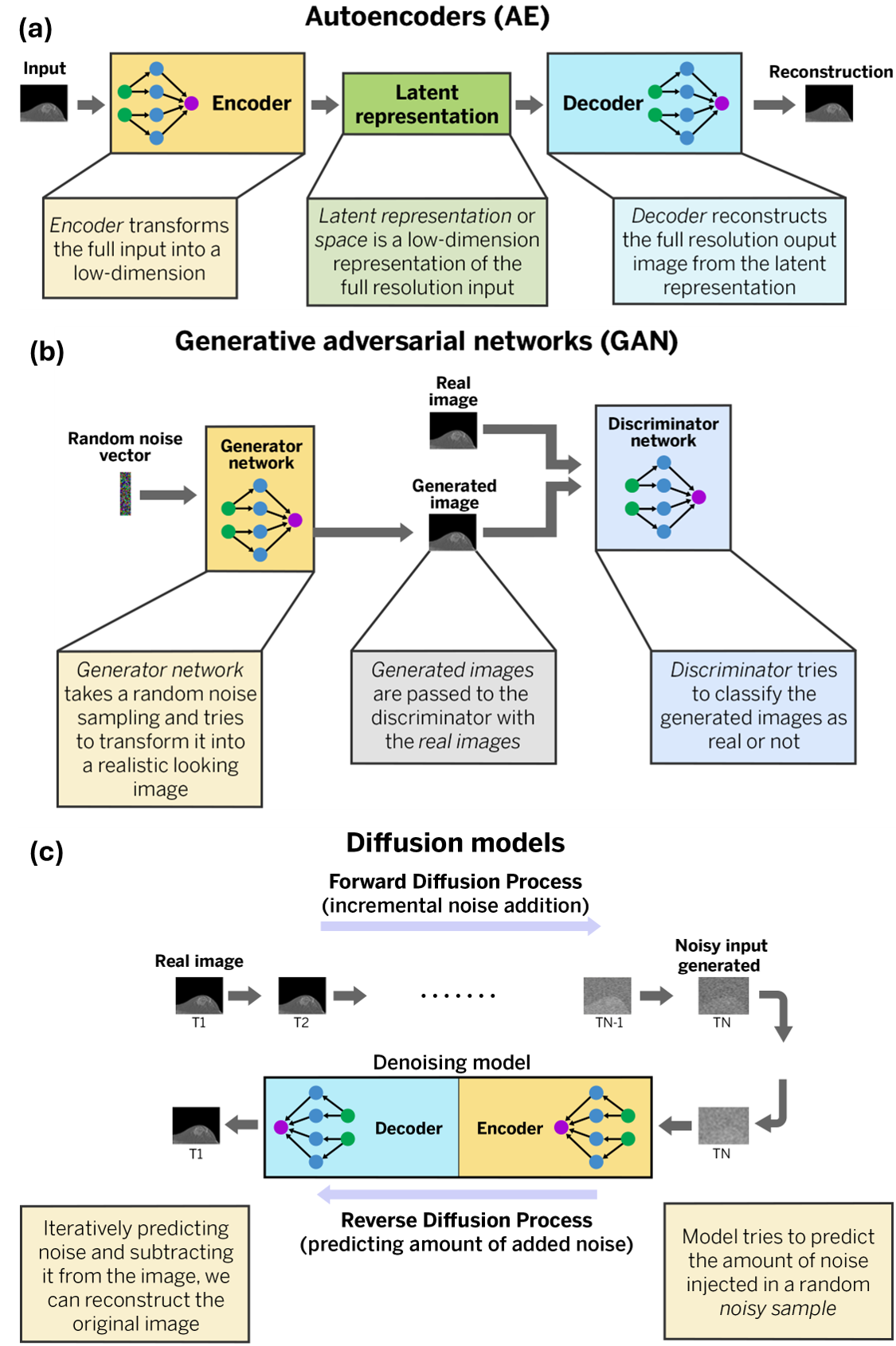}
    \caption{High-level illustrations for \textbf{(a)} Autoencoders (AE), \textbf{(b)} Generative Adversarial networks (GAN), and \textbf{(c)} Diffusion models.}
    \label{fig:structs}
\end{figure}

Image synthesis approaches can be broadly categorized into three types \cite{baraheem2023image} (an illustration of all these approaches has been shown in Figure~\ref{fig:structs}):
\begin{enumerate}
    \item \textbf{Autoencoder (AE)}: They are highly versatile, can be used as proxy feature extractors, and are able to learn robust features. At the same time, generating new samples may be difficult, tracking the latent representations is quite difficult, and they may be sensitive to hyper-parameter choice.
    \item \textbf{Generative Adversarial Network (GAN)}: They can be used to create new samples relatively easily, and the inference is very fast. However, training these models is tricky because it depends a lot on choosing the right hyper-parameters. If not done correctly, the training can fail. Additionally, it needs a lot of data and takes a long time. There is usually no way to include an objective criterion to evaluate the synthetic samples during training.
    \item \textbf{Diffusion (Diff)}: The images generated from these methods are usually very realistic (and are the current state-of-the-art), and they are simpler to optimize compared to either GANs or AEs. However, they are not very easy to comprehend, training and inference are very slow, and there is usually no way to include an objective criterion to evaluate the synthetic samples.
\end{enumerate}

The current landscape of tools that perform image synthesis is highly limited, with authors preferring to publish their respective (and usually highly specialized) pipelines \cite{nie2017medical,nie2018medical,shin2018medical,khader2022medical} instead of any general-purpose tool, or with packages specifically designed for computer vision tasks \cite{park2019semantic,rombach2022high}. The main reason for this is because computational pipelines for synthesis are highly customized (specific to the task at hand, the computational method, and other properties of the training process). MONAI Generative (MONAI-Gen) \cite{pinaya2023generative} is a library that acts as an extension built on top of MONAI \cite{cardoso2022monai} for GenAI. To the best of our knowledge, MONAI-Gen is the only cohesive software package that allows users access to a multitude of synthesis algorithms, whereas other approaches are more designed to showcase a single method or approach. However, training models using this library still requires a considerable amount of computational expertise, such as the selection of the appropriate training strategy, model topology, and other hyper-parameters. This level of expertise is usually not present in the healthcare domain. Additionally, it does not support GAN in an easy way, which could provide an important set of algorithms for healthcare (especially considering their propensity for having a built-in discriminator \cite{kazeminia2020gans,singh2021medical,xun2022generative,chen2022generative}).

In this manuscript, we present the Generally Nuanced Deep Learning Framework for Synthesis (\textbf{GaNDLF-Synth}: \url{gandlf-synth.org}), general-purpose tool for both the computational and clinical researcher interested in exploring and evaluating various image synthesis approaches for a given task. Since GaNDLF-Synth builds on top of currently available tools and libraries, including MONAI-Gen and GaNDLF-Core \cite{pati2021gandlf,pati2023gandlf} (\url{gandlf.org}), it leverage the power and understanding of the community. Specifically, these could include 1) a lot of current state-of-the-art implementations (MONAI-Gen) \cite{cardoso2022monai,pinaya2023generative}, 2) distributed compute (DeepSpeed) \cite{rasley2020deepspeed}, 3) diverse data types (GaNDLF-Core) \cite{pati2023gandlf}, 4) optimization (OpenVINO) \cite{thakur2021optimization}, 5) secure containerization (MLCube) \cite{karargyris2023federated}, and 6) direct ties with the Linux Foundation's OpenFL library for federated learning studies \cite{foley2022openfl}.

\section{Summary}

GaNDLF-Synth is designed as a higher-level abstraction supporting synthesis approaches like AE, GANs, and Diff models, enabling users to compare multiple algorithms for the same use case (Table~\ref{tab:genai_applications}) and assess their relative utility. As a Python library extension to GaNDLF, it offers a zero/low code interface, making it accessible to both computational researchers, who benefit from a unified platform for applying novel methods, and clinical researchers, who can train multiple algorithms without coding. GaNDLF-Synth is continuously updated to incorporate the latest synthesis approaches, ensuring it remains a cutting-edge tool for researchers. It can be easily installed from either Git or pip by following the installation instructions in the repository: \href{https://github.com/mlcommons/GaNDLF-Synth}{github.com/mlcommons/GaNDLF-Synth}.

\section{Discussion}

In this manuscript, we present \textbf{GaNDLF-Synth}, a zero/low-code software package that enables researchers to use GenAI methods without any software development expertise. The zero/low-code approach lowers the entry barrier and offers a high level of unified abstraction for diverse synthesis algorithm pipelines, including AE, GAN, and Diff models. This makes GenAI approachable and extensible for all researchers working in the medical imaging domain, allowing them to easily tackle multiple types of synthesis models (each with highly heterogeneous training mechanisms) for the same data.
GaNDLF-Synth is built on general-purpose tools (such as GaNDLF-Core \cite{pati2021gandlf,pati2023gandlf} and MONAI-Gen \cite{pinaya2023generative}) it has a wide range of out-of-the-box support for various data processing and different modalities, including radiology and histopathology. 


Training any synthesis model requires a great deal of computational expertise, including understanding the model topology and design, choosing the appropriate optimizer and scheduler strategies, and identifying the best metric to use for the task at hand \cite{maier2024metrics}. By ensuring a unified abstraction layer that can communicate with multiple types of synthesis approaches, GaNDLF-Synth alleviates many of these concerns regarding the design of the training pipeline, further democratizing synthesis approaches and making them more accessible to healthcare researchers. Given that synthesis approaches are highly application-specific \cite{elasri2022image,zhan2023multimodal}, providing researchers with a cohesive pipeline to train multiple approaches enables a more comprehensive analysis of different synthesis methods for their data.

We have validated this tool by leveraging publicly available datasets to train multiple synthesis models. We compared the synthesized images from each of these models with real images using first-order radiomic features. Our analyses show that the synthetic images are very similar in characteristics to the real images. The trained diffusion model approach performed the best for this specific use case, albeit with a significant inference time of around 312 hours per sample. The autoencoder model performed reasonably well and was much faster, with an inference time of around 1 second per sample.

Finally, by having seamless support for DeepSpeed \cite{rasley2020deepspeed}, GaNDLF-Synth also enables users to easily perform distributed computing i.e., the ability to utilize a wide array of heterogeneous compute capabilities (such as CPU/GPU on a single node or multiple GPUs on a single or multiple nodes). This can take the form of data parallelism \cite{li2020pytorch} (i.e., splitting the data across multiple accelerator cards, enabling faster training), or model parallelism \cite{castello2019analysis,zhuang2023optimizing} (i.e., splitting the model across multiple accelerator cards, enabling the training of larger models). This enables GaNDLF-Synth to maximize throughput without requiring additional code. Further research is needed to properly quantify the effect of such general-purpose tools on other types of data, such as 3D radiology and histopathology images, as well as non-imaging data.

\section{Resource Availability}

\subsection{Summary Statement}

    By enabling training and inference of complex GenAI methods using just text files, GaNDLF-Synth aims to democratize the use of synthesis approaches in healthcare, which has thus far required a significant amount of computational expertise. 

\subsection{Code Location}
    The GaNDLF-Synth framework showcased in this manuscript is available to use through its public repository: \href{https://github.com/mlcommons/GaNDLF-Synth}{github.com/mlcommons/GaNDLF-Synth}.

\subsection{Potential Use Cases}
    
    There are multiple applications of GenAI in healthcare, and some of them are highlighted in Table~\ref{tab:genai_applications}. Essentially, GenAI can be used to generate realistic healthcare imaging data based on a population cohort. The use cases of such techniques can be towards preservation of privacy (where a model for a downstream task is trained on the ``synthetic'' data as opposed to ``real'' patient data) as well as to data augmentation (where a model for a downstream task can be augmented with new data to improve its performance). Additionally, as most of the generative models are trained in an unsupervised manner, such models are capable of learning general data distribution properties (such as generating missing parts of images \cite{kofler2023brain}, or for images affected by motion artefacts \cite{xue2012motion}). The pre-trained model blocks can be then fine-tuned for specific tasks (segmentation, classification). This approach is beneficial, especially in the scenario when the data for downstream tasks is scarce.

\subsection{Licensing}
    The license used is \href{https://www.apache.org/licenses/LICENSE-2.0}{Apache 2.0}.

\subsection{Ethical Considerations}
    Fully de-identified and anonymized datasets were used to perform validation of this tool. 

%
The work follows appropriate ethical standards in conducting research and writing the manuscript, following all applicable laws and regulations regarding treatment of animals or human subjects.

\section{Methods}

\subsection{Software Stack and Flowchart}

\begin{figure}
    \centering
    \includegraphics[width=0.975\textwidth]{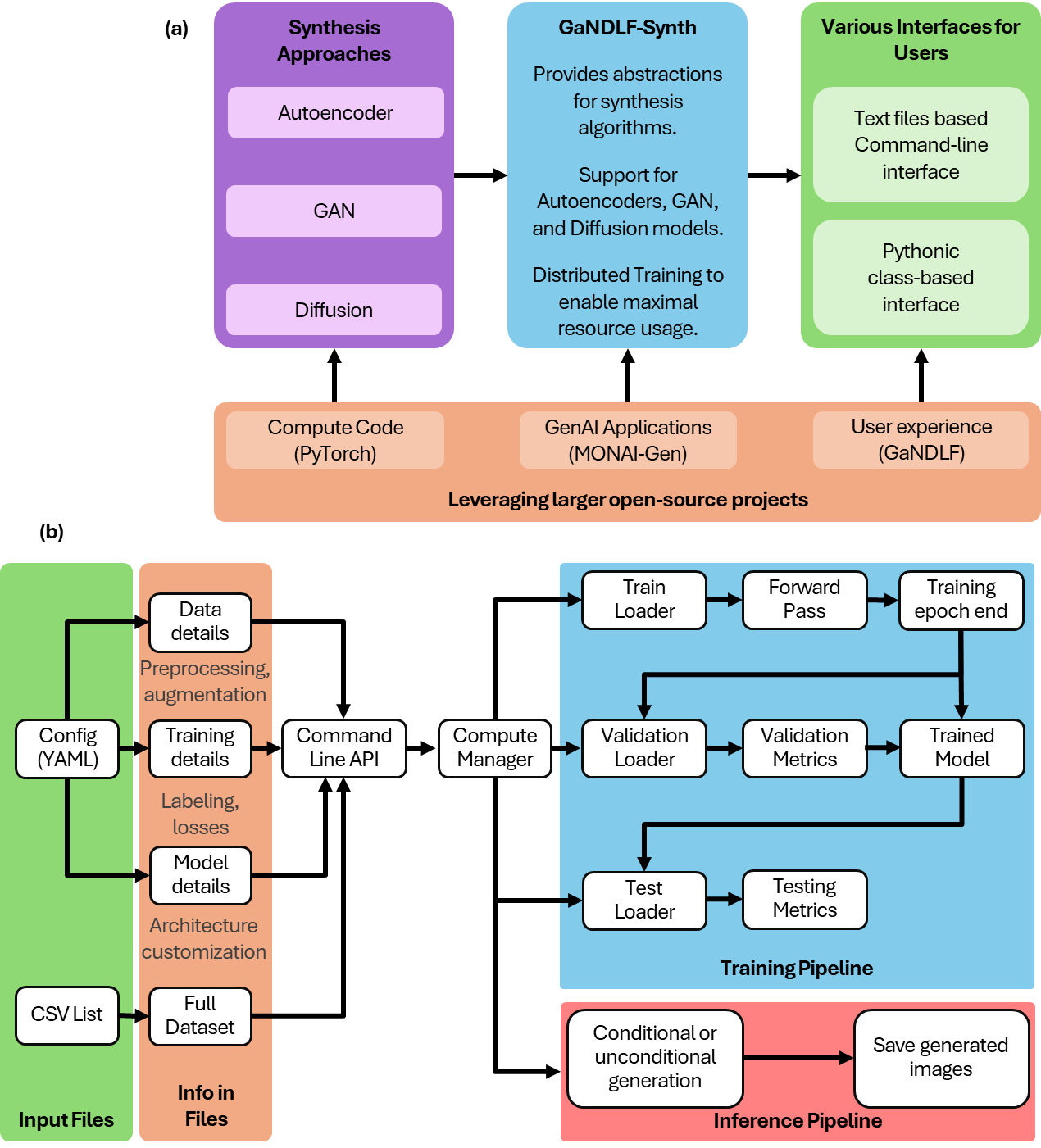}
        \caption{Illustration showcasing the internal components of \textbf{GaNDLF-Syth}. \textbf{(a)} is showcasing the software stack, with special focus on the usage of existing functionalities from PyTorch and GaNDLF for compute functionality and user experience, and leveraging a customizable interface to allow easy algorithmic extensions. \textbf{(b)} is showcasing the flowchart depicting the overall training and inference procedure offered in GaNDLF-Synth.}
    \label{fig:stack}
    \label{fig:stack_abstract}
    \label{fig:flowchart}
\end{figure}

\textbf{GaNDLF-Synth} is built on well-known public libraries, such as PyTorch \cite{paszke2019pytorch}, Lightning \cite{falcon2019pytorch}, MONAI-Gen \cite{pinaya2023generative}, and SimpleITK \cite{lowekamp2013design}, ensuring the long-term sustainability of the project dependencies. By having GaNDLF \cite{pati2021gandlf,pati2023gandlf} as a core dependency, GaNDLF-Synth inherits the majority of the user experience, including the ability to interact with the entire codebase using just text-based YAML files. The data used for training and inference is passed through CSV files. This combination lowers the barrier for users to start gaining experience with synthesis-based models while allowing computational experts to inject reproducibility into their synthesis routines. A high-level illustration of the stack is shown in Figure~\ref{fig:stack}.a.



\subsection{Functionalities}

\textbf{GaNDLF-Synth} adopts a zero/low code approach pioneered through its core dependency with GaNDLF \cite{pati2023gandlf}, allowing users to configure experiments via YAML files (i.e., structured text files) that are automatically parsed and validated. This removes extensive programming effort on the user side while providing a large degree of flexibility when needed. For example, GaNDLF-Synth streamlines the construction of the data loader (including pre-processing and augmentation routines) by only requiring users to set up the data directory according to the chosen labeling paradigm, and constructing a simple text-based CSV file which provides the input data and the appropriate targets.

GaNDLF-Synth provides a highly customizable abstraction layer for model design which addresses the heterogeneous nature of their various computing pipelines (e.g., AE approaches might need a different set of losses compared to GANs, and Diff models would need iterative noise removal). Additionally, it currently provides end users with a wide library of out-of-the-box models that can be easily used for real-world use cases focused on synthesis. It also offers an easy way to organize and structure code which allows developers and advanced users to include their custom solutions with little overhead. It also enables scalability of experiments by natively supporting distributed training and inference strategies, such as DeepSpeed \cite{rasley2020deepspeed} and data distributed parallel. 

GaNDLF-Synth supports both unconditional and conditional generation, enabling the creation of models in fully unconditional and conditional setups. It automatically handles output logging, enabling effective metric tracking, and supports highly customizable checkpointing, which is crucial for fault tolerance and portability of trained models. The checkpoints are fully compatible with the Lightning framework format. Additionally, GaNDLF-Synth integrates with popular MLOps \cite{kreuzberger2023machine} tools for experiment tracking, both externally hosted like WanDB and locally deployed like MLFlow \cite{alla2021beginning}. An illustration of the workflow is shown in Figure~\ref{fig:flowchart}.b.

\section{Validation}

\subsection{Code Reproducibility}

All training and inference processes are based on text-based configurations, which can be easily versioned using Git. Samples of various  configurations are provided in the repository to facilitate reproducibility and transparency. Additionally, the codebase is comprehensively covered with multiple continuous integration (CI)/continuous deployment (CD) pipelines through GitHub Actions \cite{kinsman2021software,saroar2023developers}. Each pull request undergoes rigorous evaluation against existing code coverage and unit tests, ensuring the robustness and reliability of the code.

\label{sec:validation}


\coi{We declare that we do not have any conflicts of interest.}

\acks{Research reported in this publication was partly supported by the Informatics Technology for Cancer Research (ITCR) program of the National Cancer Institute (NCI) of the National Institutes of Health (NIH) under award numbers U01CA242871 and U24CA279629. The content of this publication is solely the responsibility of the authors and does not represent the official views of the NIH.

This research was supported in part by Lilly Endowment, Inc., through its support for the Indiana University Pervasive Technology Institute. 

We gratefully acknowledge Polish high-performance computing infrastructure PLGrid (HPC Center: ACK Cyfronet AGH) for providing computer facilities and support within computational grant no. PLG/2024/017091.}

\bibliography{_references}

\end{document}